\documentclass[10pt,twocolumn,letterpaper]{article}

\usepackage{3dv}
\usepackage{times}
\usepackage{epsfig}
\usepackage{graphicx}
\usepackage{amsmath}
\usepackage{amssymb}
\usepackage{subcaption}
\graphicspath{{images/}}

\usepackage[utf8]{inputenc}
\usepackage{siunitx}
\usepackage{enumitem}
\usepackage{booktabs}
\usepackage{nopageno}

\usepackage[pagebackref=true,breaklinks=true,letterpaper=true,colorlinks,bookmarks=false]{hyperref}

\newcommand{\djt}[1]{}
\newcommand{\fede}[1]{}
\newcommand{\husi}[1]{}
\newcommand{\todo}[1]{}

\newcommand{\figref}[1]{Fig.~\ref{#1}}
\newcommand{\secref}[1]{Sec.~\ref{#1}}
\newcommand{\tabref}[1]{Table~\ref{#1}}
\newcommand{\best}[1]{\textbf{#1}}

\newcommand{\rec}{y}
\newcommand{\pred}{y_x}
\newcommand{\predgt}{y_t}
\newcommand{\gt}{t}

\newcommand{\generator}{\mathcal{G}}
\newcommand{\encoderdepth}{\mathcal{E}_\text{dep}}
\newcommand{\encodervoxel}{\mathcal{E}_\text{vox}}
\newcommand{\latdepth}{l_{\text{dep}}}
\newcommand{\latvoxel}{l_{\text{vox}}}

\newcommand{\Dvox}{\mathcal{D}_\text{vox}}
\newcommand{\Dlat}{\mathcal{D}_l}

\newcommand{\Lpred}{\mathcal{L}_{x \rightarrow y} (\encoderdepth, \generator)}
\newcommand{\Lpredgt}{\mathcal{L}_{t \rightarrow y} (\encodervoxel, \generator)}
\newcommand{\Lgany}{\mathcal{L}_{\text{GAN-}y}}
\newcommand{\Lganl}{\mathcal{L}_{\text{GAN-}l}}

\threedvfinalcopy

\def\threedvPaperID{191} 

\ifthreedvfinal\fi
\begin{document}

\title{Adversarial Semantic Scene Completion from a Single Depth Image}

\author{Yida Wang, David Joseph Tan, Nassir Navab, Federico Tombari\\
Technische Universität München\\
Boltzmannstraße 3, 85748 Garching bei München\\
}

\maketitle

\begin{abstract}
We propose a method to reconstruct, complete and semantically label a 3D scene from a single input depth image. We improve the accuracy of the regressed semantic 3D maps by a novel architecture based on adversarial learning. In particular, we suggest using multiple adversarial loss terms that not only enforce realistic outputs with respect to the ground truth, but also an effective embedding of the internal features. This is done by correlating the latent features of the encoder working on partial 2.5D data with the latent features extracted from a variational 3D auto-encoder trained to reconstruct the complete semantic scene. 
In addition, differently from other approaches that operate entirely through 3D convolutions, at test time we retain the original 2.5D structure of the input during downsampling to improve the effectiveness of the internal representation of our model. We test our approach on the main benchmark datasets for semantic scene completion to qualitatively and quantitatively assess the effectiveness of our proposal. 
\end{abstract}

\section{Introduction}

Inspired by the way humans can imagine the structure of a room by looking at an image, we propose an algorithm that reconstructs the entire scene geometry and semantics from a single depth image.
By directly reconstructing the scene from one view, the challenge is to plausibly complete the scene in place of the hidden structures that are not visible from the input depth image. 
To this end, we utilize a learning strategy that allows the algorithm to simultaneously perceive the objects in the scene and use its contextual shape to fill the hidden structures.
In addition, we simultaneously estimate a semantic segmentation of the completed 3D scene geometry.
%

\begin{figure}[t]
	\centering
	\includegraphics[width=1\linewidth]{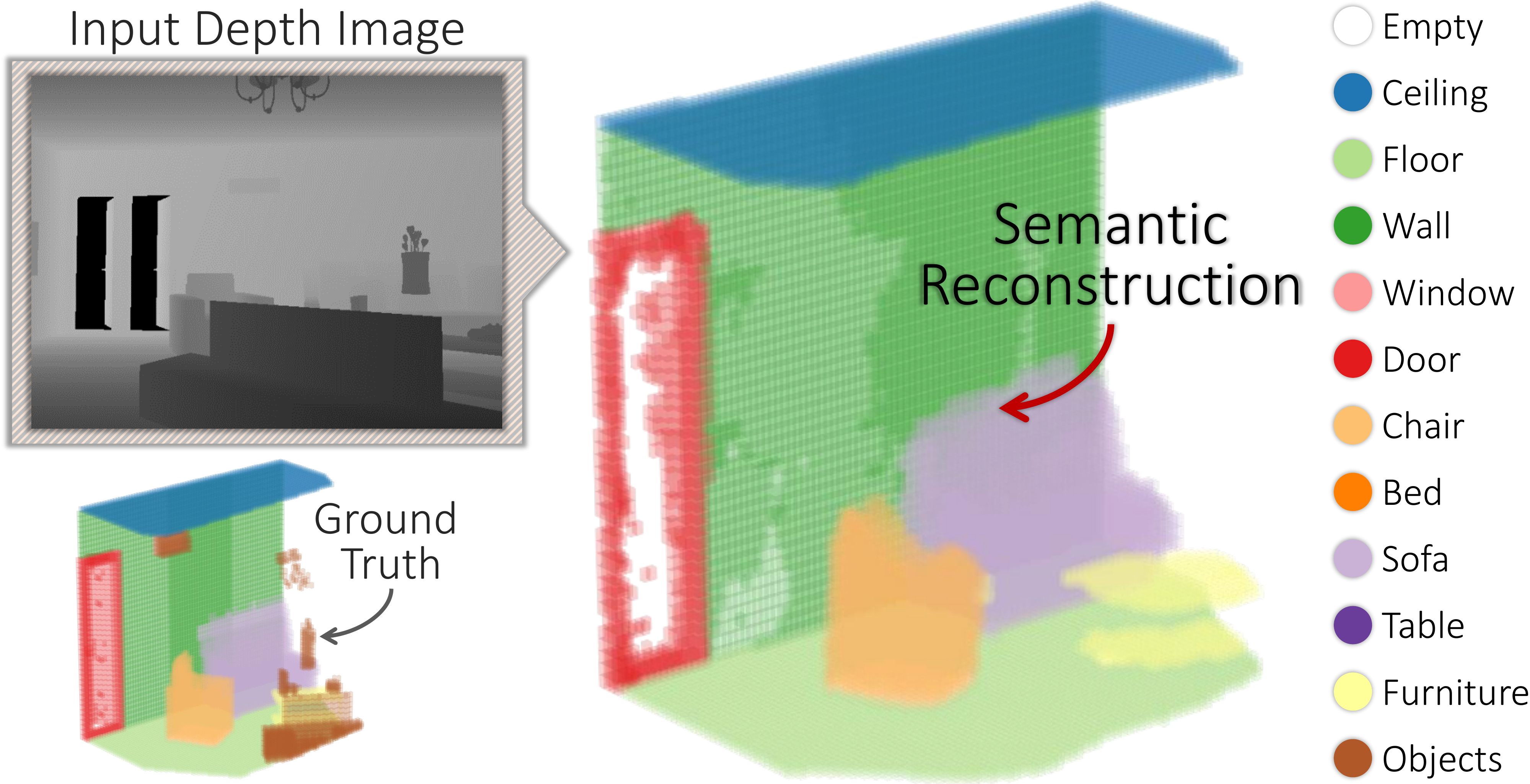}
	\caption{The input depth image and the output semantic 3D reconstruction. \label{fig:teaser}}
\end{figure}

Reconstructing the environmental information in 3D space from a single viewpoint is relevant for a lot of tasks in the field of augmented reality~\cite{van2010survey}, robotic perception~\cite{kober2013reinforcement} and scene understanding~\cite{li2009towards}, where users and autonomous agents often have only a limited set of observations of the surrounding, and would benefit from a complete semantic reconstruction of the scene geometry. 
To push the scientific effort along these directions, recently large-scale benchmark datasets, such as SUNCG~\cite{song2016ssc}, NYU~\cite{silberman2012indoor}, ScanNet~\cite{dai2017scannet} and SceneNet~\cite{handa2015scenenet}, have been proposed to evaluate different visual scene understanding tasks including those of scene completion and semantic segmentation.

A few methods have recently been proposed in the direction of 3D shape completion. 
In particular, SSCNet~\cite{song2016ssc} demonstrated good results in the joint task of scene completion and semantic segmentation by means of a CNN~\cite{SzegedyLJSRAEVR14, he2015deep}\@. 
They encode the depth image into volumetric space using the Truncated Signed Distance Function (TSDF) from KinectFusion~\cite{newcombe2011kinectfusion}.
Differently, 3D-RecGAN++~\cite{yang20183d} suggests using an adversarial approach to learn how to realistically complete partial object shapes from common classes. 
Generative models have also been proposed to generate 3D data directly from 2D images such as the 3D Inductor~\cite{gadelha20163d}.

In this work, we focus on the data acquired from depth cameras, with the goal of reconstructing and semantically labeling the whole scene from one single range image. 
As a scene may contain small objects and complicated shapes, we apply a generative adversarial model for this semantic completion task.
Combined with an encoder and a generator, our architecture uses depth images directly as the input information and generate 3D volumetric data whose elements are labeled with object categories.
Specifically, we use two discriminators to train the architecture to back-project the depth information into the 3D volumetric space with semantic labels. 
One discriminator is used to optimize the entire architecture by comparing the reconstructed semantic scene with the ground truth.
%
%
Since the 3D variational auto-encoders models the latent features of the volumetric data very well, we designed our architecture such that our encoder for depth images learns similar latent features. 
To do so, we introduce another discriminator for optimizing the learnt latent features.

To summarize, we have two main contributions. 
Firstly, we propose the first generative adversarial network aimed at semantic 3D scene completion, and we demonstrate how the adversarial approach is a meaningful choice for the task at hand. Secondly, we enforce adversarial learning not just on the output reconstruction, but also on the latent space to improve the quality of the results. We evaluate our approach on the main benchmark dataset for semantic scene completion to qualitatively and quantitatively assess the effectiveness of our proposal. 

\begin{figure*}[!ht]
	\centering
	\includegraphics[width=1\linewidth]{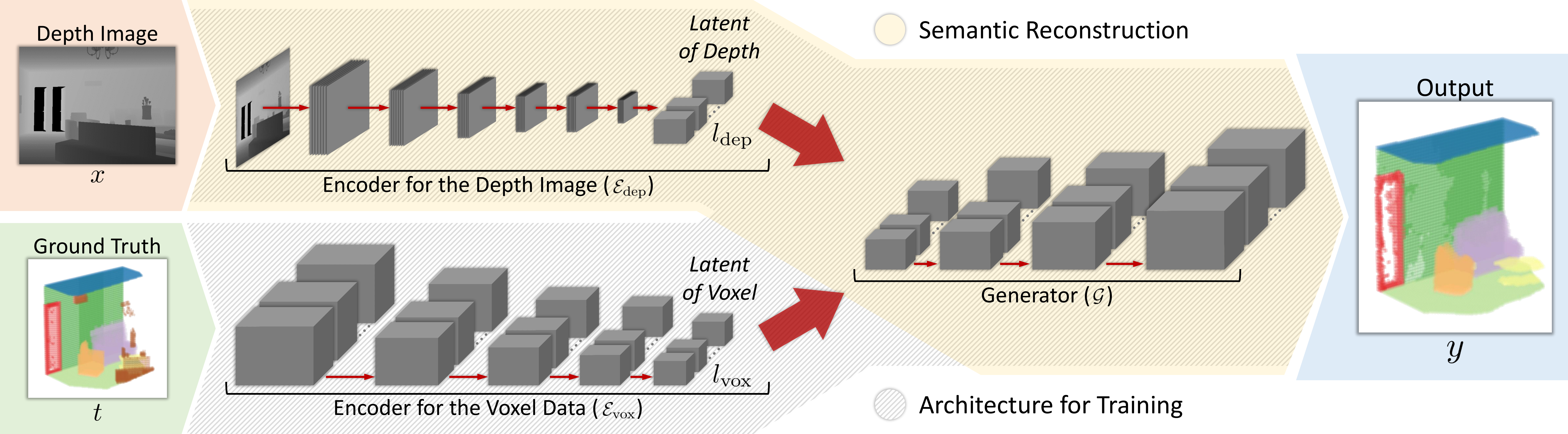}
	\caption{Deep architecture for the semantic reconstruction in \secref{sec:reconstruction} and for the training procedure in \secref{sec:training}. The former is the concatenated architecture of the encoder $\encoderdepth$ and the generator $\generator$ reconstructs from the depth image to a voxel data while the latter is the concatenated architecture of $\encodervoxel$ and $\generator$ is a 3D variational auto-encoder~\cite{brock2016generative} for self-reconstruction.}
	\label{fig:architecture}
\end{figure*}

\section{Related work}

Being particularly difficult and training intensive, the task of shape completion from 2.5D has only, with the recent explosion of deep learning, started to become a main research trend in the community. 
SSCNet~\cite{song2016ssc} proposes a CNN-based architecture that carries out jointly the 3D scene completion and the semantic labeling from a single depth image. A voxel-wise softmax loss function is proposed as the optimizer for learning semantic segmentation of volumetric elements.
For training, the method assumes to know the viewpoint as well as the alignment of the depth maps and the reconstructed volumes to a common 3D reference frame. Differently, our approach drops such assumptions and can work without the information regarding the camera pose or the global alignment. 
On a different task, 3D-RecGAN++~\cite{yang20183d} suggests learning a 3D adversarial generative model to complete partial 3D shapes of common object classes. The use of the adversarial loss is motivated to provide realistic and plausible interpolations of the missing shape parts. 

Scene and object completion has been investigated also from RGB data. MarrNet~\cite{marrnet} proposes to reconstruct 3D object from 2.5D sketches with normal, depth and silhouette information extracted from 2D images. Inspired by MarrNet, the encoder of our architecture is mainly composed of 2D convolutional operators while the generator is mainly composed of 3D deconvolutional operators. The difference lies in the fact that the latent variables of our model are learned to be similar to the feature extracted from a 3D VAE trained on the complete volumetric data.

PointOutNet~\cite{fan2017point} proposes an encoder-decoder deep architecture to complete 3D objects from RGB images in the form of 3D coordinates.
3D-$R^{2}N^{2}$~\cite{choy20163d} tries to reconstruct a volumetric representation of an object from an RGB image by training a recurrent neural network over a latent representation of the RGB data. 
In addition, by combining scene reconstruction and GAN, 3D-Scene-GAN~\cite{yu20183d} is introduced for reconstructing complicated 3D scenes from RGB views with mesh and texture by applying a discriminator to distinguish between the rendered 2D images of the scene and real ones.

On a different topic, feature representations for generative models has been often deployed for reconstruction tasks, \eg by means of Variational Auto-Encoders (VAE)~\cite{blei2017variational,kingma2013auto} and conditional VAE (CVAE)~\cite{Kingma2014Semi,NIPS2015_5775}, which are two popular methods to learn features from an input data in continuous latent spaces trained via variational inference.
3D VAE~\cite{brock2016generative} is also introduced by replacing 2D convolutional kernels with 3D kernels for auto-encoding voxel data.

\section{Semantic reconstruction}
\label{sec:reconstruction}

The semantic reconstruction algorithm takes a single view of the scene, depicted by a depth image $x$, to predict its 3D volumetric representation $\rec$.
The voxels of $\rec$ are semantically labeled with $N_c$ object classes, denoted as an $N_c\times1$ one-hot vector, \ie a binary vector where one of its element has a value of $1$ to indicate the object category while the other elements remain zero. 
%
Considering that the image has a limited view of the entire scene, constrained by the sensor's viewpoint, the objective of our deep learning approach is also to complete the scene by revealing the hidden structures that are not visible in the input. 
Therefore, simulateneously learning the geometric structure and the semantic information allows the algorithm to learn the contextual cues that can in turn represent the objects in the reconstruction.

Specifically, the depth image is a $640\times480$ image that represents the $z$-axis of the camera coordinate system. As input to our deep learning architecture, this image is down-sampled to $320\times240$ in order to conserve GPU memory. 
The resulting volumetric reconstruction is represented by $N_c$ grids of size $40\times80\times80$ filled with binary elements presenting the labels for each of the $N_c$ objects. For simplicity, we denote this 4D data as $40\times80\times80\times N_c$.

From the depth image to the 3D volume, our architecture is a concatenation of an encoder $\encoderdepth$ with 2D convolutional operators that convert the input depth image into a lower-dimensional latent feature $\latdepth$; and, a generator $\generator$ with 3D deconvolutional kernels that takes $\latdepth$ to build the semantic reconstruction.
This architecture is illustrated in \figref{fig:architecture}. 
%


\paragraph{Encoder for depth image.} 

The encoder $\encoderdepth$ compresses the depth image into a feature in the latent space. 
Its architecture is a concatenated network that sequentially combines 2D convolutional layers and max-pooling layers. 
The operators for the paired convolutional and pooling layers are 2D convolutional kernels with, respectively, the size of $3\times3$ and stride of $1\times1$ and the size of $2\times2$  with stride of $2\times2$.
Each of these paired layers is processed by a leaky ReLU activation function~\cite{nair2010rectified}. 
Therefore, the output of every ReLU activation is a multi-channel 2D image. After six convolutions operations, the result is an 80-channel $5\times3$ image which is reshaped into a set of 3D volume of size $5\times3\times5\times16$. The output of the encoder represents the latent feature $\latdepth$ of the semantic reconstruction architecture.

\paragraph{Generator.} 

With the goal of regressing the semantic reconstruction, the generator $\generator$ unwraps the latent feature to a higher dimensional voxel data.
We assemble the generator with 3D deconvolutional layers with the size of $3\times3\times3$ and stride of $2\times2\times2$ which are processed by the ReLU function as activation.
After four deconvolutional layers, the output of the generator is the voxel-wise classification $y$. 
By doing this, $y$ is presented in the shape of $80\times48\times80\times N_c$.

%


\section{Architecture for training}
\label{sec:training}

Although our semantic reconstruction algorithm in \secref{sec:reconstruction} could be optimized only with encoder $\encoderdepth$ and generator $\generator$, the performance after training this way is subpar (see \secref{sec:exp_suncg}).
Hence, we include three components during the training process to improve the performance -- 
(1)~the encoder for the voxel data,
(2)~the discriminator for the reconstruction 
and 
(3)~the discriminator for the latent features.

\begin{figure}[t]
	\centering
	\includegraphics[width=1\linewidth]{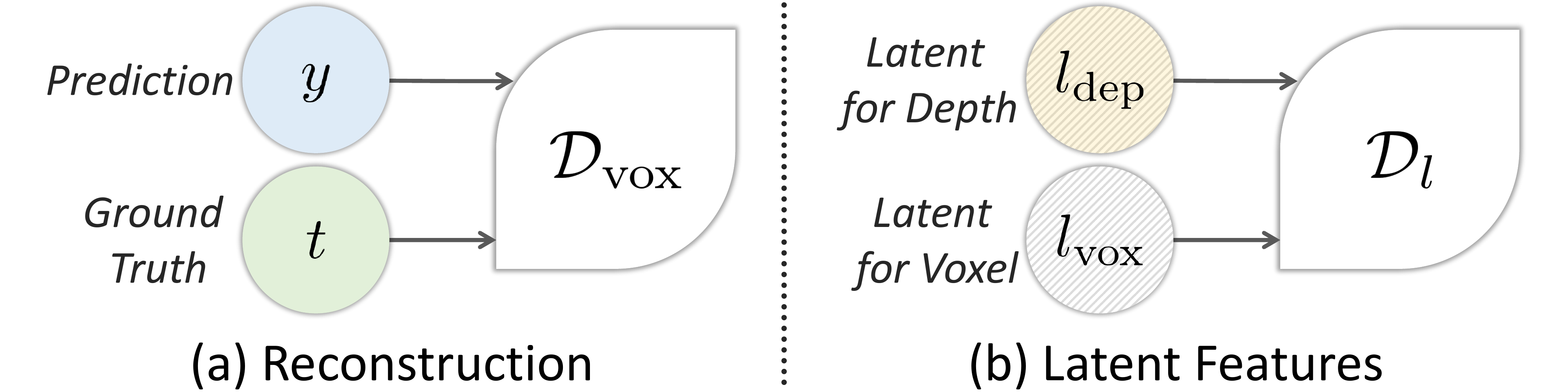}
	\caption{Variables associated to each discriminator.}
	\label{fig:discriminators}
\end{figure}

Specifically, we introduce another encoder $\encodervoxel$ to extract the feature $\latvoxel$ such that the latent feature from the encoder $\encoderdepth$ is driven to be similar to a feature extracted from $\encodervoxel$. 
Thus, a discriminator $\Dlat$ is used to optimize this similarity as illustrated in \figref{fig:discriminators} and consequently updates the parameters in $\encoderdepth$.
Notably, $\encodervoxel$ is optimized together with the generator $\generator$ as a 3D variational auto-encoder (3D VAE)~\cite{brock2016generative}  to learn meaningful weights from training samples representing complete 3D semantic volumes.

\begin{figure*}[t]
	\centering
	\includegraphics[width=1\linewidth]{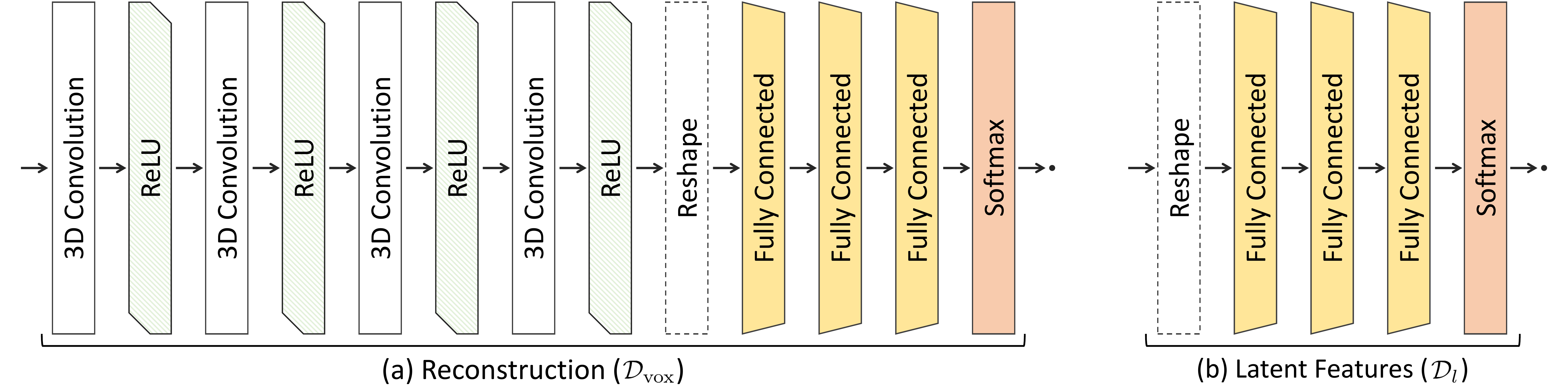}
	\caption{Architecture of the two discriminators.}
	\label{fig:d_architecture}
\end{figure*}

\paragraph{Encoder for the voxel data.}

Since reconstructing from one image has a restrictive view of the scene, we want to make the latent features $\latdepth$, extracted from the depth image, to be similar to the complete volumetric data in order to incorporate structures that are not visible from the input.
We introduce another encoder $\encodervoxel$ to extract the feature $\latvoxel$ into the architecture for learning. 
The input of $\encodervoxel$ is the ground truth volumentric data with semantic labels such that all the operators in its architecure are 3D convolution kernels with the size of $3\times3\times3$ and stride of $2\times2\times2$ as shown in \figref{fig:architecture}.
The last layer of the encoder $\encodervoxel$ produces 16 blocks. 
The size of the output from $\encodervoxel$ is set to be the same as $\encoderdepth$ because we want to make the latent representation of the input depth images to be as similar as possible to that of the ground truth volumetric representations.
Thus, measuring the similarity between $\latdepth$ and $\latvoxel$ is possible because the latent representation compresses the results for both the depth and voxel data. 
%
As illustrated in \figref{fig:architecture},
the latent features from both the encoders $\encoderdepth$ and $\encodervoxel$ go through the same generator $\generator$ to predict semantic volumetric data.
%

\paragraph{Discriminator for the reconstruction.} 
Encouraged by the benefits of the generative models trained with adversarial techniques~\cite{creswell2018generative}, 
we introduce the discriminator $\Dvox$ in training to optimize our semantic reconstruction by comparing our prediction against the ground truth as shown in \figref{fig:discriminators}. 
The architecture of $\Dvox$ is similar to the encoder $\encodervoxel$ except for the last layer. 
In \figref{fig:d_architecture} (a), all the four 3D convolutions have $3\times3\times3$ kernels with stride of $2\times2\times2$.
Then, the output of the last convolutional layer with the size of $5\times3\times5\times16$ is reshaped to a vector of 1200 dimensions. 
This is processed by three fully-connected layers with output sizes, respectively, of 256, 128 and 1.
Hence, the final logit is a binary indicator to determine whether the predicted volumetric data is the expected ones or not, which is widely used in GAN~\cite{rosca2017variational}.

\paragraph{Discriminator for the latent features.}
Since the output of $\encoderdepth$ and $\encodervoxel$ are passed to the same generator $\generator$, the resulting latent feature from the depth image $\latdepth$ is driven to be similar to the feature extracted from the ground truth volumetric data $\latvoxel$. 
We introduce another discriminator $\Dlat$ aiming at distinguishing the latent descriptors illustrated in \figref{fig:discriminators} and consequently updating the parameters in $\encodervoxel$.
As input to $\Dlat$, the latent variables are reshaped from $5\times3\times5\times16$ to a vector of 1200 dimensions. 
The architecture of $\Dlat$ in \figref{fig:d_architecture} (b) is constructed purely by three fully-connected layers with output sizes of 256, 128 and 1. 
Finally, the output of the discriminator is also a logit where 1 indicates that the latent feature from the depth image is similar to the feature of the 3D VAE\@; otherwise, the value is zero.

\section{Optimization}
\label{sec:optimization}

The goal of the optimization is to enforce the latent features from the depth image ($\latdepth$) and the predicted reconstruction ($\rec$) to resemble the latent features of the 3D VAE ($\latvoxel$) and the ground truth volumetric data ($\gt$), respectively. Since the architecture for the semantic reconstruction and the 3D VAE share the same generator (see \figref{fig:architecture}), we distinguish their results by denoting $\pred$ as the prediction from the semantic reconstruction while $\predgt$ from the 3D VAE\@.
 
\paragraph{Loss functions.}
When we solely consider the semantic reconstruction architecture (see \secref{sec:reconstruction}), the loss that compares the prediction and the ground truth for all the $N_c$ objects is represented as 
\begin{align}
	\label{equ:loss_gen}
	\Lpred = \sum_{c=1}^{N_c}\left[ \epsilon(\pred(c),\gt(c)) \right]
\end{align}
where we define the per-object error as
\begin{align}
	\epsilon(q,r) = -\gamma r \log q - (1-\gamma)(1 - r)\log(1 - q)
\end{align}
with $\gamma$ as the hyper-parameter which weighs the relative importance of false positives against false negatives. 
Consequently, the error penalizes when the prediction and the ground truth are distinct.

To further improve the reconstruction performance using GAN (see \secref{sec:training}), we include an adversarial loss   
\begin{align}
	\label{equ:loss_gan_voxel_gen}
	\Lgany(\encoderdepth,\generator) = -\log(\Dvox(\pred))
\end{align}
based on the trained discriminator $\Dvox$ that optimizes the semantic reconstruction architecture by updating the parameters of its encoder and generator. 
On the other hand, training for the parameters in $\Dvox$ entails a loss function 
\begin{align}
	\label{equ:loss_gan_voxel_dis}
	\Lgany(\Dvox) = -\log(\Dvox(\gt)) - \log(1 - \Dvox(\pred))
\end{align}
so that the discriminator $\Dvox$ could be further optimized to be capable of distinguishing the generated volumetric data from the ground truth.

As for the 3D VAE, we can train this architecture by minimizing a loss similar to~\eqref{equ:loss_gen}. However, since we use the ground truth reconstruction as the input, the loss function
\begin{align}
	\label{equ:loss_gen_latent}
	\Lpredgt = \sum_{c=1}^{N_c}\left[ \epsilon(\predgt(c),\gt(c)) \right]
\end{align}
enforces the predicted reconstruction $\predgt$ to be similar to its input.
By training with variational inference by optimizing the evidence lower bound (ELBO)~\cite{blei2017variational,kingma2013auto}, the latent variables are distributed in a simple Gaussian distribution.

In reference to semantic reconstruction architecture, the 3D VAE influences the latent variable $\latdepth$ to be as similar to $\latvoxel$ as possible by using discriminator $\Dlat$ to determine whether $\latdepth$ is presented similar to $\latvoxel$. 
Therefore, similar to $\Dvox$, optimizing the similarity between the latent features uses another discriminator $\Dlat$ such that the loss function to update the encoder $\encoderdepth$ is   
\begin{align}
	\label{equ:loss_gan_latent_enc}
	\Lganl(\encoderdepth) = -\log(\Dlat(\latdepth)) 
\end{align}
while training for $\Dlat$ involves 
\begin{align}
	\label{equ:loss_gan_latent_dis}
	\Lganl(\Dlat) = -\log(\Dlat(\latvoxel)) - \log(1 - \Dlat(\latdepth))~.
\end{align}

\paragraph{Minimization.}
Now that we have all the loss functions, our optimization is defined as a combination of five components. 
The first two are based on the architecture for training in \figref{fig:architecture}. From the depth image $x$ and the ground truth $t$, we separately train them one after the other for the samples in a mini-batch with   
\begin{align}
	&\min(\Lpred)~\text{and}\\
	&\min(\Lpredgt) 
\end{align}
so that the parameters of the architectures are updated alternatively. 
At the same time, the variational inference sets a constraint on the latent variables as a Gaussian distribution which makes it easier for the output of both of the encoders to match with each other. 

Assuming that the discriminators are trained, we can fix their parametric model in order to update the encoder to move towards
\begin{align}
	\label{equ:opt_encoder}
	\min(\Lganl(\encoderdepth)) 
\end{align}
while update both the encoder and the generator toward
\begin{align}
	\label{equ:opt_generator}
	\min(\Lgany(\encoderdepth,\generator))
\end{align}
which are also optimized alternatively.

Finally, the two discriminators are trained by minimizing 
\begin{align}
	\label{equ:opt_discriminator_y}
	&\min(\Lgany(\Dvox))~\text{and}\\
	\label{equ:opt_discriminator_l}
	&\min(\Lganl(\Dlat))
\end{align}
such that the former is used to penalize poorly reconstructed voxel data in reference to the ground truth while the latter makes the latent codes computed from the depth image similar to the latent feature extracted from a well trained 3D VAE\@.
Notably, the discriminators are updated when the accuracy in distinguishing the generated outputs are lower than specific level~\cite{creswell2018generative}.
We set this threshold to 15\% in our experiments.

In practice, we use the Adam optimizer~\cite{kingma2014adam} 
with a learning rate of 0.0001.

\section{Implementation details}

We use the paired depth image and semantically labeled volumes provided by SUNCG~\cite{song2016ssc} and NYU~\cite{silberman2012indoor}.
The size of volumetric data with the object labels is $240\times240\times240\times N_c$ where $N_c$ is set to $12$. 
Due to the limited GPU memory, we down-sample the data to $80\times48\times80\times N_c$ by max-pooling with $3\times3\times3$ kernel and $3\times3\times3$ stride.
In this manner, the original volumetric data is presented in a space with a lower resolution which is suitable for training in a single GPU with no more than 12 GB memory.
In our experiments, we use a single NVIDIA TITAN Xp for training and the batch size is set to be 8. 
The depth images are also resized from $640\times480$ to $320\times240$ with a bilinear interpolation.

The 12 object classes in our experiments are based on SUNCG~\cite{song2016ssc} that includes: empty space, ceiling, floor, wall, window, door, chair, bed, sofa, table, furniture and small objects.
Since the ratios of samples in each categories are not balanced, we redesign the evaluation strategy in \secref{sec:exp} to concentrate on reconstructing important objects in the indoor condition with small amount of voxels such as furnitures and small objects.


\section{Experiments}
\label{sec:exp}

\begin{table*}[!ht]
\centering
\resizebox{\textwidth}{!}
{\begin{tabular}{l|cccccccccccc|c}
	\toprule	
		\multicolumn{1}{c}{} &
		empty & ceil. & floor & wall & win. & door & chair & bed & sofa & table & furn. & objs. & 
		\emph{Avg.} \\
	\midrule
	3D VAE~\cite{brock2016generative} & 49.3 & 26.1 & 33.2 & 29.7 & 14.4 & 4.6 & 0.7 & 16.4 & 13.9 & 0.0 & 0.0 & 0.0 & 30.8 \\
	3D-RecGAN++~\cite{yang20183d} & 49.3 & 32.6 & 37.7 & 36.0 & 23.6 & 13.6 & 8.7 & 20.3 & 16.7 & 9.6 & 0.2 & 3.6 & 36.1 \\
	Ours without $\Dlat$ & 49.6 & 42.0 & 35.9 & 44.8 & 28.5 & 25.5 & 15.4 & 28.6 & 20.1 & 21.5 & 11.5 & 6.5 & 42.7 \\
	Ours without $\Dvox$ & 49.6 & 39.0 & 35.7 & 43.4 & 26.8 & 23.8 & 18.5 & 29.2 & 22.4 & 16.8 & 10.4 & 5.3 & 41.7 \\
	Ours (\emph{Proposed}) & 49.7 & 41.4 & 37.7 & 45.8 & 26.5 & 26.4 & 21.8 & 25.4 & 23.7 & 20.1 & 16.2 & 5.7 & \best{44.1} \\
	\bottomrule
	\end{tabular}
}
\caption{Semantic scene completion results on the SUNCG test set with depth map for IoU in percentage.\label{tab:suncg_iou}}
\end{table*}

\begin{table*}[!ht]
\centering
\resizebox{\textwidth}{!}
{\begin{tabular}{l|cccccccccccc|c}
	\toprule	
		\multicolumn{1}{c}{} &
		empty & ceil. & floor & wall & win. & door & chair & bed & sofa & table & furn. & objs. & 
		\emph{Avg.} \\
	\midrule
	3D VAE~\cite{brock2016generative} & 99.6 & 18.8 & 68.9 & 63.6 & 25.0 & 8.5 & 4.2 & 16.4 & 9.5 & 1.3 & 0.4 & 2.6 & 65.6 \\
	3D-RecGAN++~\cite{yang20183d} & 99.9 & 21.5 & 76.2 & 78.8 & 31.9 & 15.3 & 8.1 & 18.7 & 10.2 & 2.9 & 1.4 & 4.3 & 79.4 \\
	{Ours without $\Dlat$} & 100.0 & 29.1 & 72.8 & 92.9 & 29.7 & 20.2 & 9.9 & 20.8 & 13.5 & 2.6 & 6.2 & 3.0 & 92.3 \\
	{Ours without $\Dvox$} & 99.9 & 28.6 & 70.3 & 91.5 & 28.3 & 18.8 & 9.1 & 20.2 & 12.7 & 2.6 & 4.9 & 2.6 & 90.1 \\
	Ours (\emph{Proposed}) & 100.0 & 29.1 & 76.2 & 94.2 & 32.0 & 22.7 & 11.4 & 21.9 & 14.2 & 3.1 & 7.6 & 3.6 & \best{94.5} \\
	\bottomrule
	\end{tabular}
}
\caption{Semantic scene completion results on the SUNCG test set with depth map for mAP in percentage.\label{tab:suncg_map}}
\end{table*}


We evaluated on the SUNCG dataset~\cite{song2016ssc} that includes pairs of depth images and the corresponding semantically labelled 3D reconstructions.

\paragraph{Evaluation Strategy.}

Considering that this dataset is for the indoor environments, over 90\% of the reconstructed scene is empty. 
Then, when we exclude the empty spaces, simple structures such as the wall, floor and ceiling dominate the voxels in the scene. 
This means that the ratio of the number of voxels for different object classes is not balanced. 
For instance, we noticed that the SUNCG test sets~\cite{song2016ssc} do not have enough small objects and furnitures.
In this case, if the learned architecture enhances its ability to predict the empty spaces and the simple structures, their accuracy is significantly higher than the results predicted by an architecture that focuses on distinguishing the other object classes. 

Since the ratio of voxels for small objects and furnitures in the training dataset are higher than the one in the test set in SUNCG~\cite{song2016ssc}, we design a 10-fold cross validation by splitting the training data which was introduced by~\cite{huang2007labeled}. 
The entire dataset is divided into ten folds with the same amount of samples, the evaluation procedure then uses 1 of the 10 folds as the test set and the remaining 9 as the training dataset. 
Thereafter, the final result is the average of the ten evaluations.

\paragraph{Metric.}

We evaluate the performance of the reconstructor based on the intersection over union (IoU) and the mean average precision (mAP) of the predicted voxel labels compared to ground truth labels~\cite{song2016ssc} where we evaluate the IoU of each object classes on both the observed and occluded voxels for semantic scene completion.
Notably, instead of taking the average IoU and mAP as the mean of the results from individual categories, we calculate the average with respect to the number of voxels in each category.

\paragraph{Comparison.}

We compare our results against 
3D VAE~\cite{brock2016generative} and 3D-RecGAN++~\cite{yang20183d}.
In order to directly estimate the volumetric reconstruction solely from the input depth image, 
we modify~\cite{brock2016generative, yang20183d} by scaling the surface generated by the depth image through bilinear interpolation to fit the $80\times48\times80$ volumetric grid which serves as the input to~\cite{brock2016generative, yang20183d}. 
Furthermore, we added the loss function from~\eqref{equ:loss_gen} in training to perform semantic segmentation.
Notably, the U-Net~\cite{ronneberger2015u} connection between encoder and decoder in 3D-RecGAN++~\cite{yang20183d} are still applied by resizing the scale of every layers.
In addition, we further investigate the advantage of the discriminators by evaluating our approach without $\Dlat$ and $\Dvox$.
Based on \secref{sec:optimization}, when implementing our approach without $\Dlat$,~\eqref{equ:opt_encoder} and~\eqref{equ:opt_discriminator_l} are discarded in the optimization; while, the implementation without $\Dvox$ discards~\eqref{equ:opt_generator} and~\eqref{equ:opt_discriminator_y}.

\begin{table*}[t]
\centering
\resizebox{\textwidth}{!}
{\begin{tabular}{l|cccccccccccc|c}
	\toprule	
		\multicolumn{1}{c}{} &
		empty & ceil. & floor & wall & win. & door & chair & bed & sofa & table & furn. & objs. & 
		\emph{Avg.} \\
	\midrule
	3D VAE~\cite{brock2016generative} & 49.4 & 33.3 & 25.3 & 32.4 & 16.9 & 9.3 & 5.6 & 19.2 & 14.7 & 1.1 & 0.0 & 0.0 & 31.5 \\
	3D-RecGAN++~\cite{yang20183d} & 49.6 & 35.1 & 31.8 & 39.2 & 23.7 & 17.9 & 11.5 & 26.1 & 22.6 & 18.1 & 5.1 & 3.0 & 37.7 \\
	Ours without $\Dlat$ & 49.6 & 42.4 & 35.8 & 44.4 & 29.2 & 24.8 & 17.2 & 30.6 & 24.2 & 19.5 & 11.5 & 4.4 & 42.4 \\
	Ours without $\Dvox$ & 49.7 & 43.9 & 37.3 & 45.9 & 26.7 & 29.2 & 20.1 & 24.0 &24.6 & 26.1 & 19.8 & 9.0 & 44.3 \\
	Ours (\emph{Proposed}) & 49.8 & 49.6 & 42.7 & 51.2 & 24.2 & 34.9 & 23.0 & 28.1 & 30.4 & 29.9 & 22.0 & 11.5 & \best{51.4} \\
	\bottomrule
	\end{tabular}
}
	\caption{Semantic scene completion results finetuned on the NYU training set with real world depth map for IoU in percentage.\label{tab:nyu_iou}} 
\end{table*}

\begin{table*}[t]
\centering
\resizebox{\textwidth}{!}
{\begin{tabular}{l|cccccccccccc|c}
	\toprule	
		\multicolumn{1}{c}{} &
		empty & ceil. & floor & wall & win. & door & chair & bed & sofa & table & furn. & objs. & 
		\emph{Avg.} \\
	\midrule
	3D VAE~\cite{brock2016generative} & 99.8 & 25.0 & 53.8 & 70.9 & 19.3 & 7.4 & 4.2 & 14.3 & 9.4 & 1.1 & 1.2 & 0.9 & 68.4 \\
	3D-RecGAN++~\cite{yang20183d} & 99.9 & 27.3 & 67.5 & 87.6 & 27.0 & 15.8 & 8.0 & 19.2 & 12.0 & 2.2 & 3.4 & 1.8 & 86.5 \\
	Ours without $\Dlat$ & 100.0 & 28.9 & 72.1 & 92.7 & 29.6 & 19.8 & 9.9 & 20.8 & 13.3 & 2.7 & 6.6 & 2.9 & 91.9 \\
	Ours without $\Dvox$ & 100.0 & 29.2 & 76.8 & 94.5 & 31.9 & 22.6 & 11.5 & 21.9 & 14.2 & 3.2 & 8.2 & 4.1 & 94.8 \\
	Ours (\emph{Proposed}) & 100.0 & 30.8 & 79.1 & 96.6 & 35.4 & 26.9 & 17.0 & 13.9 & 15.8 & 3.6 & 9.7 & 5.5 & \best{97.2} \\
	\bottomrule
	\end{tabular}
}
	\caption{Semantic scene completion results finetuned on the NYU training set with real world depth map for mAP in percentage.\label{tab:nyu_map}} 
\end{table*}

\begin{figure*}[t]
	\centering
	\includegraphics[width=\linewidth]{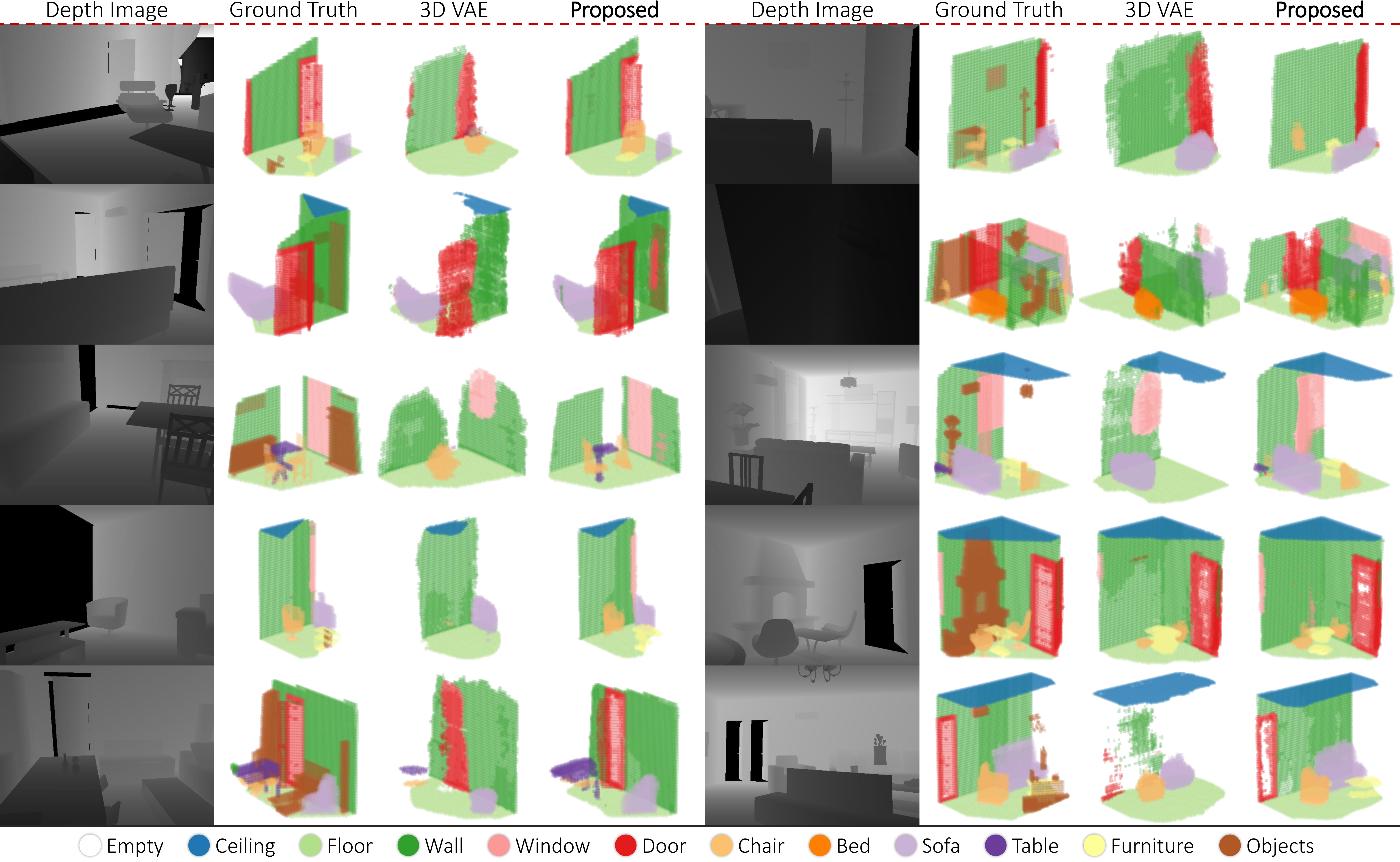}
	\caption{GAN for semantic 3D reconstruction from depth images.\label{fig:recon_suncg}}
\end{figure*}

\subsection{SUNCG}
\label{sec:exp_suncg}

SUNCG~\cite{song2016ssc} is a dataset of 3D scenes which contains pairs of depth image and its corresponding volumetric scene where all objects in the scene are semantically annotated.
We implemented the 10-fold validation on the pairs for the 111,697 different scenes.

\paragraph{Comparison against other approaches.}

The evaluation on both the IoU and mAP in \tabref{tab:suncg_iou} and \tabref{tab:suncg_map} shows that our generative model performs better than 3D VAE~\cite{brock2016generative} and 3D-RecGAN++~\cite{yang20183d} which are the recent works on 3D generative architectures. 
We acquired an IoU of 44.1\% and an mAP of 94.5\% that is 8\% and 15.1\% better than the next best performing approach.

\paragraph{Comparison on the architecture for learning.}

To understand the advantage of incorporating the components in learning, we investigate learning our method without the discriminators. 
Without the discriminator for the latent features, our performance decreases by 1.4\% in IoU  and 2.2\% in mAP\@; while, without the discriminator for the reconstruction, the results decrease by 2.4\% in IoU and 4.4\% in mAPo\@.
However, it is noteworthy to mention that, even without these discriminators, our method still achieves better results compared to both 3D VAE~\cite{brock2016generative} and 3D-RecGAN++~\cite{yang20183d}.

\paragraph{Performance on smaller objects.}

If we look closely on \tabref{tab:suncg_iou}, our approach has a significant improvement over 3D VAE~\cite{brock2016generative} and 3D-RecGAN++~\cite{yang20183d} on smaller objects like the class of table, furniture and objects wherein~\cite{brock2016generative} produced an IoU of zero. 
The reason behind this improvement is because the adversarial training is especially helpful in reconstructing and completing small objects compared to 3D VAE~\cite{brock2016generative}.
Note that these results are also validated by evaluating the mAP in \tabref{tab:suncg_map}.

Since the latent space is continouos, this implies that it reserves regions for the object classes with a smaller amount of voxels in the scene or a fewer samples in the training dataset. 
Therefore, while all methods can reconstruct the common objects such as the ceiling, floor and walls with correct labels as illustrated in both \tabref{tab:suncg_iou} and \tabref{tab:suncg_map}, the main advantage of our work is the capacity to reconstruct and classify every type of object labels.

\paragraph{Qualitative results.}

We illustrate the qualitative results in \figref{fig:recon_suncg} and compare them with 3D VAE~\cite{brock2016generative} and the ground truth. Based on these voxel representations, we can clearly visualize the superiority of our algorithm to reconstruct more detailed structures compared to~\cite{brock2016generative}.
Therefore, this confirms the advantage of our approach to reconstruct not only the larger structures but also the smaller objects in the scene.

\subsection{Fine-tune with NYU}
\label{sec:exp_nyu}

The objective of this section is to investigates whether an increase in the size of the learning dataset from a different source can improve the performance of the algorithm or confuse the learned model.

In this section, we include the NYU dataset~\cite{silberman2012indoor} which is also an indoor scene dataset. It contains both the depth images captured by Kinect and the 3D models. 
This includes the volumetric 3D data with the annotated object labels for every voxels in 1,449 scenes.
The semantic annotations for the volumetric data in this dataset consist of 33 objects in 7 categories.
Note that, due to the limited amount of 1,449 volumetric scenes from the NYU dataset, this size is insufficient to learn a deep learning architecture. 
Thus, we only use the NYU to supplement our training dataset while testing on SUNCG for the 12 categories.
This requires us to relabel the object classes of the volumetric data in NYU to match the labels provided by SUNCG dataset.

\paragraph{Comparison against other approaches.}
Similar to \secref{sec:exp_suncg}, this procedure is implemented on all the five approaches that we are comparing. 
While fine-tuning with the NYU dataset, our experiments show that the combination of the two datasets improve the perfomance of our algorithm. 
From \tabref{tab:suncg_iou} to \tabref{tab:nyu_iou} and \tabref{tab:suncg_map} to \tabref{tab:nyu_map}, we experience an increase in IoU by 7.3\% and in mAP by 2.7\%. 
Although both the 3D VAE~\cite{brock2016generative} and 3D-RecGAN++~\cite{yang20183d} also experienced an increase in performance, the difference is not significant which counts for a maximum of 1.6\% increase in IoU. 
Note that, in \tabref{tab:nyu_iou}, the results on smaller objects for 3D VAE~\cite{brock2016generative} remains close to zero or zero.

\paragraph{Comparison on the architecture for learning.}
When we learn our architecture with the discriminators, the effect of the improvement is negligible. Without the discriminator for the latent features, the IoU even decreased from 42.7\% to 42.4\%; while, without the discriminator for the reconstruction, the IoU increases only from 41.7\% to 44.3\%.
Therefore, based on this experiment, we can attribute the significant improvement of our work's perfomance to the discriminators in the training architecture.

\section{Conclusion}

We have proposed a novel approach for semantic scene completion from a single depth map, which exploits the power of adversarial training to regress accurate reconstructions without the need of additional assumptions or the camera pose information. 
Our proposal relies on the enforcement of two adversarial losses -- one aimed at making the output realistc; while, the other aimed at imitating the embedding learned via auto-encoder from the complete volumetric data. We have demonstrated the effectiveness of our approach on a reference benchmark dataset such as SUNCG\@. 
%
The future work aims at modifying our architecture to overcome the memory limitation so to process higher resolution samples, this allowing a direct comparison with approaches such as SSCNet~\cite{song2016ssc}.

{\small
\bibliographystyle{ieee}
\bibliography{egbib}
}

\end{document}